\documentclass[runningheads]{llncs}
\usepackage{llncsdoc}
\usepackage{url}
\usepackage[numbers]{natbib}
\usepackage{times}
\usepackage{latexsym}
\usepackage{graphicx}
\usepackage{subfigure}
\usepackage{color}
\usepackage[normalem]{ulem}
\usepackage{multirow}
\usepackage{amsmath}
\usepackage{amssymb}
\usepackage{amsfonts}

\usepackage{algorithm}
\usepackage{algorithmic}

\renewcommand{\bibsection}{\section*{References}}

\newcommand{\af}[1]{\textcolor{black}{#1}}
\newcommand{\zsz}[1]{\textcolor{black}{#1}}

\newcommand{\dhlee}[1]{\textcolor{black}{#1}}

\begin{document}
\title{Difference Target Propagation}
\titlerunning{Difference Target Propagation}

\author{Dong-Hyun Lee\inst{1} \and Saizheng Zhang\inst{1} \and Asja Fischer\inst{1} \and\\
Yoshua Bengio\inst{1,2}}
\authorrunning{D-H. Lee, S. Zhang, A. Fischer and Y. Bengio}
\institute{
Universit\'e de Montr\'eal, Quebec, Canada
\and
CIFAR Senior Fellow
}

\maketitle

\begin{abstract}
Back-propagation has been the workhorse of recent successes of deep learning
but it relies on infinitesimal effects (partial derivatives) in order to perform credit
assignment. This could become a serious issue as one considers deeper and more
non-linear functions, e.g., consider the extreme case of non-linearity where the relation
between parameters and cost is actually discrete. Inspired by the biological
implausibility of back-propagation, a few approaches have been proposed in the
past that could play a similar credit assignment role. In this spirit,
we explore a novel approach to credit assignment in deep networks that we call
target propagation. The main idea is to compute targets rather than gradients, at
each layer. Like gradients, they are propagated backwards. In a way that is related
but different from previously proposed proxies for back-propagation which
rely on a backwards network with symmetric weights, target propagation relies
on auto-encoders at each layer. Unlike back-propagation, it can be applied even
when units exchange stochastic bits rather than real numbers. We show that a linear
correction for the imperfectness of the auto-encoders, called difference
target propagation, is very effective to make
target propagation actually work, leading to results comparable to back-propagation
for deep networks with discrete and continuous units and denoising auto-encoders
and achieving state of the art for stochastic networks. \footnote{This paper was accepted in ECML/PKDD 2015. Please cite like following :
Dong-Hyun Lee, Saizheng Zhang, Asja Fischer and Yoshua Bengio, Difference Target Propagation, in Machine Learning and Knowledge Discovery in Databases, pages 498-515, Springer International Publishing, 2015}

\end{abstract}

\section{Introduction}

Recently, deep neural networks have achieved great success in hard AI tasks ~\citep{Bengio-2009-book,Hinton-et-al-2012,Krizhevsky-2012-small,Sutskever-et-al-arxiv2014},
mostly relying on back-propagation  as the main 
way of performing credit assignment over the different sets of 
parameters associated with each layer of a deep net.  Back-propagation exploits the chain rule of derivatives in order to convert a loss gradient on the activations over layer $l$ (or time $t$, for recurrent nets) into a loss gradient on the activations over layer $l-1$ (respectively, time $t-1$).  However, as we consider deeper networks-- e.g., consider the recent best ImageNet competition entrants ~\citep{Szegedy-et-al-arxiv2014} with 19 or 22 layers --
longer-term dependencies, or stronger non-linearities, the composition of
many non-linear operations becomes more strongly non-linear. To make this concrete,
consider the composition of many hyperbolic tangent units. In general, this
means that derivatives obtained by back-propagation are becoming either very small
(most of the time) or very large (in a few places). In the extreme (very deep
computations), one would get discrete functions, whose derivatives are 0
almost everywhere, and infinite where the function changes discretely.
Clearly, back-propagation would fail in that regime. In addition,
from the point of view
of low-energy hardware implementation, the ability to train deep networks
whose units only communicate via bits would also be interesting.

This limitation \af{of} back-propagation to working with precise derivatives and smooth
networks is the main machine learning motivation for this paper's exploration
into an alternative principle for credit assignment in deep
networks. Another motivation arises from the lack of biological
plausibility of back-propagation, for the following reasons: (1) the
back-propagation computation is purely linear, whereas biological neurons
interleave linear and non-linear operations, (2) if the feedback paths were
used to propagate credit assignment by back-propagation, they would need precise
knowledge of the derivatives of the non-linearities at the operating point
used in the corresponding feedforward computation, (3) similarly, these
feedback paths would have to use exact symmetric weights (with the same
connectivity, transposed) of the feedforward connections, (4) real neurons
communicate by (possibly stochastic) binary values (spikes), (5) the
computation would have to be precisely clocked to alternate between feedforward
and back-propagation phases, and (6) it is not clear where the output
targets would come from. 

The main idea of target propagation is to associate with each feedforward
unit's activation value a {\em target value} rather than a {\em loss
  gradient}. The target value is meant to be close to the activation value
while being likely to have provided a smaller loss (if that value had been
obtained in the feedforward phase). In the limit where the target is very
close to the feedforward value, target propagation should behave like
back-propagation. This link was nicely made
in ~\citep{LeCun-dsbo86,Lecun-these87}, which introduced the idea of target
propagation and connected it to back-propagation via a Lagrange multipliers
formulation (where the constraints require the output of one layer to equal
the input of the next layer). A similar idea was recently proposed where
the constraints are relaxed into penalties, yielding a different
(iterative) way to optimize deep networks ~\citep{Carreira-Perpinan-and-Wang-AISTATS2014}. Once
a good target is computed, a layer-local training criterion can be defined
to update each layer separately, e.g., via the delta-rule (gradient
descent update with respect to the cross-entropy loss).

By its nature, target propagation can in principle handle stronger
(and even discrete) non-linearities, and it deals with biological
plausibility issues \af{(1), (2), (3) and (4)}  described above. Extensions of the
precise scheme proposed here could handle \af{(5) and (6) as well}, but this is
left for future work.

In this paper, \af{we describe how the general idea of target propagation
by using auto-encoders to assign targets to each layer
 (as introduced in an earlier technical report~\citep{Bengio-arxiv2014}) can be 
employed for supervised training of deep neural networks (section \ref{sec:FromTargets} and \ref{sec:howToAssign}).
We continue by introducing a linear correction for the imperfectness of the auto-encoders (\ref{sec:DiffTargetProp}) leading to robust training in practice. Furthermore, we show how the same principles can be applied to replace back-propagation in the training of auto-encoders (section \ref{sec:diff-tg-ae}).  In section \ref{sec:experiments}} we provide several experimental results on rather deep neural networks as well as discrete and stochastic networks \af{and auto-encoders}. The results show that the
proposed form of target propagation is comparable to
back-propagation with RMSprop~\citep{tieleman2012lecture} 
- a very popular setting to train deep networks nowadays- \af{and
achieves state of the art for training  stochastic neural nets on MNIST.}

\section{\af{Target Propagation}}\label{sec:targetProp}

Although many variants of the general principle of target propagation can be devised, this paper focuses on a specific approach, \af{which is based on the
ideas presented in an earlier technical report~\citep{Bengio-arxiv2014} and is
described in the following.}


\subsection{Formulating Targets}\label{sec:FromTargets}

Let us consider an ordinary \af{(supervised) deep network learning process,
where the training data is drawn from an unknown data distribution $p(\mathbf{x}, \mathbf{y})$.}
The network structure is defined by
\begin{equation}
\mathbf{h}_i = f_i(\mathbf{h}_{i-1}) = s_i( W_i \mathbf{h}_{i-1}), \enspace i=1,\dots,M
\end{equation}
\af{where $\mathbf{h}_i$ is the state of the $i$-th hidden layer (where $\mathbf{h}_M$ corresponds to the output of the network and  $\mathbf{h}_0= \mathbf{x}$)
and $f_i$ is the $i$-th layer feed-forward mapping, defined by 
a non-linear activation function $s_i$  (e.g. the hyperbolic tangents or the sigmoid function) and the weights $W_i$ of the $i$-th layer.
Here, for simplicity of notation, the bias term of the $i$-th layer is included in $W_i$.
We refer to 
the subset of network parameters defining the 
mapping between the $i$-th and the $j$-th layer ($0\leq i<j \leq M$) as}
$\mathbf{\theta}_W^{i,j} = \{W_k, k=i+1,\dots,j\}$. 
\af{Using} this notion, we \af{can write} $\mathbf{h}_j$ as a function of $\mathbf{h}_i$ \af{depending on parameters $\mathbf{\theta}_W^{i,j}$, that is we can write}  $\mathbf{h}_j=\mathbf{h}_j(\mathbf{h}_i;\mathbf{\theta}_W^{i,j})$.

\af{Given a sample $(\mathbf{x}, \mathbf{y})$,  let $L(\mathbf{h}_M(\mathbf{x};\mathbf{\theta}_W^{0,M}), \mathbf{y})$ be an arbitrary global loss function measuring 
the appropriateness of the network output $\mathbf{h}_M(\mathbf{x};\mathbf{\theta}_W^{0,M})$ 
for the target $\mathbf{y}$,
e.g. the MSE or cross-entropy for binomial random variables.
 Then, the training objective corresponds to adapting the network parameters $\mathbf{\theta}_W^{0,M}$ so as to minimize the expected global loss  $\mathop{\mathbb{E}}_{p} \{L(\mathbf{h}_M(\mathbf{x};\mathbf{\theta}_W^{0,M})\mathbf{y})\}$ under the data distribution $p(\mathbf{x}, \mathbf{y})$.
 For $i=1,\dots,M-1$ we can write
\begin{eqnarray}
L(\mathbf{h}_M(\mathbf{x};\mathbf{\theta}_W^{0,M}), \mathbf{y})
=L(\mathbf{h}_M(\mathbf{h}_i(\mathbf{x};\mathbf{\theta}_W^{0,i});\mathbf{\theta}_W^{i,M}), \mathbf{y}) \enspace  
\end{eqnarray}
to emphasize the dependency of the loss on the state of the $i$-th layer.}
 
Training a network with back-propagation corresponds to propagating error
signals through the network \af{to calculate the derivatives of the global loss with respect to the parameters of each layer. Thus, the error} signals indicate how the 
parameters of the network should be updated to decrease the expected loss. 
However, in  very deep networks with strong non-linearities, error propagation could become useless  in lower layers due to 
exploding or vanishing gradients, as explained above.  

\af{To avoid this problems, the basic idea of target propagation is to assign to each $\mathbf{h}_i(\mathbf{x};\mathbf{\theta}_W^{0,i})$ a  nearby value $\hat{\mathbf{h}}_i$ which (hopefully) leads to a lower global loss, that is which has the objective to fulfill}
\begin{equation}\label{eq:target_property}
L(\mathbf{h}_M(\hat{\mathbf{h}}_i;\mathbf{\theta}_W^{i,M}), \mathbf{y}) < L(\mathbf{h}_M(\mathbf{h}_i(\mathbf{x};\mathbf{\theta}_W^{0,i});\mathbf{\theta}_W^{i,M}), \mathbf{y}) \enspace.
\end{equation}
\af{Such a $\hat{\mathbf{h}}_i$ is called a \textit{target}  for the $i$-th layer.  }

\af{Given a target $\hat{\mathbf{h}}_i$  we now would like to change the network parameters to make $\mathbf{h}_i$ move a small step towards $\hat{\mathbf{h}}_i$, since -- if the path leading from
$\mathbf{h}_i$ to $\hat{\mathbf{h}}_i$ is smooth enough -- we would expect to yield  a decrease of the global loss.
 To obtain an update direction for $W_i$ based on  $\hat{\mathbf{h}}_i$ we can define a layer-local target loss $L_i$, for example by using the MSE}
\begin{equation}
L_i (\hat{\mathbf{h}}_i, \mathbf{h}_i) = ||\hat{\mathbf{h}}_i - \mathbf{h}_i(\mathbf{x};\mathbf{\theta}_W^{0,i})||^ 2_2 \enspace.
\end{equation}
\af{Then, $W_i$ can be updated}
locally within its layer via stochastic gradient descent, 
where $\hat{\mathbf{h}}_i$ is considered as a {\em constant}
with respect to $W_i$. \af{That is}
\begin{equation}
W_i^{(t+1)} = W_i^{(t)} - \eta_{f_i} \frac {\partial L_i(\hat{\mathbf{h}}_i, \mathbf{h}_i)} {\partial W_i} = 
W_i^{(t)} - \eta_{f_i} \frac {\partial L_i(\hat{\mathbf{h}}_i, \mathbf{h}_i)} {\partial \mathbf{h}_i} \frac{\partial \mathbf{h}_i(\mathbf{x};\mathbf{\theta}_W^{0,i})} {\partial W_i} \enspace ,
\end{equation}
\af{where $\eta_{f_i}$ is a layer-specific learning rate.}

\af{Note, that in this context, derivatives can be used without 
difficulty, because they correspond to computations performed 
inside a single layer. Whereas, the problems with the severe non-linearities observed for back-propagation arise when the chain rule is applied through many layers}. This motivates
target propagation methods to serve as alternative credit assignment in the context of a composition of many non-linearities.

\af{However, it is not directly clear how to compute a target that guarantees a decrease of the global loss (that is how to compute a $\hat{\mathbf{h}}_{i}$ for which equation \eqref{eq:target_property} holds) or that at least leads to a decrease of the  local loss $L_{i+1}$ of  the next  layer, that is 
\begin{equation}\label{eq:target_property_local}
L_i( \hat{\mathbf{h}}_{i+1}, f_i(\hat{\mathbf{h}}_{i}) ) < 
 L_i( \hat{\mathbf{h}}_{i+1}, f_i(\mathbf{h}_{i}) ) \enspace.
\end{equation}}
Proposing and validating answers to this question is the subject of the rest
of this paper.

\subsection{How to assign a proper target to each layer}\label{sec:howToAssign}

\af{Clearly, in a supervised learning setting, the top layer target should be directly driven from the gradient of 
the global loss
\begin{equation}\label{eq:target_toplayer}
\hat{\mathbf{h}}_{M} = \mathbf{h}_{M} - \hat\eta \frac { \partial L (\mathbf{h}_{M}, \mathbf{y}) } { \partial \mathbf{h}_{M} } \enspace,
\end{equation}
where $\hat \eta $ is usually a small step size.   Note, that if we use the MSE as global loss and $\hat\eta=0.5$ we get $\hat{\mathbf{h}}_{M}= \mathbf{y}$.}

\begin{figure}[ht]
\vspace{-10pt}
\begin{center}
\centerline{\hspace{1cm}\includegraphics[width=0.4\columnwidth]{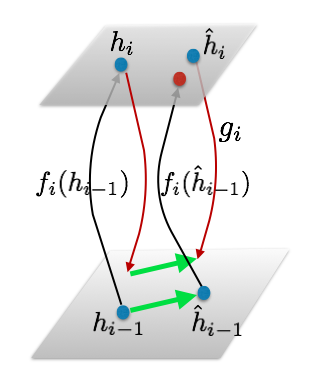} \hspace{1cm}
\includegraphics[width=0.5\columnwidth]{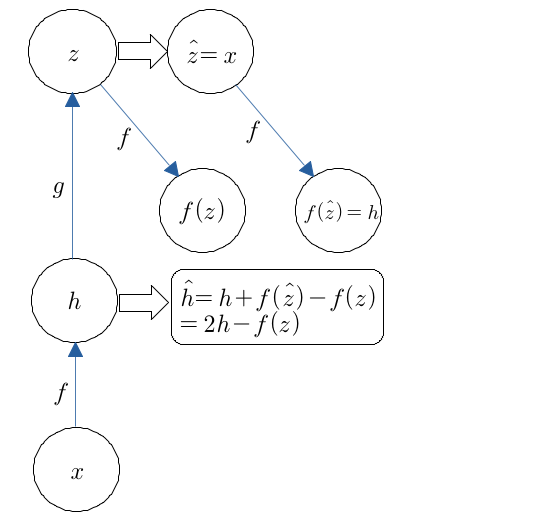}}
\caption{(left) How to compute a target in the lower layer via difference target propagation. $f_i(\hat{\mathbf{h}}_{i-1})$ should be closer to $\hat{\mathbf{h}}_{i}$ than $f_i(\mathbf{h}_{i-1})$.
(right) Diagram of the back-propagation-free auto-encoder via difference target propagation.}
\label{fig:BFAE-diagram}
\end{center}
\vspace{-30pt}
\end{figure}

\af{But how can we define targets for the intermediate layers?
In the previous technical report \citep{Bengio-arxiv2014}}, it was suggested
\af{to take advantage of an ``approximate inverse''.
To formalize this idea, suppose that for each $f_i$ we have a function $g_i$ such that 
\begin{equation}
f_i(g_i({\mathbf{h}_i})) \approx {\mathbf{h}_i} \enspace \,\,\,\, \text{or} \enspace \,\,\,\,  g_i(f_i({\mathbf{h}_{i-1}})) \approx {\mathbf{h}_{i-1}} \enspace.
\end{equation}}
Then, choosing 
\begin{equation}\label{eq:tp}
\hat {\mathbf{h}}_{i-1} = g_i(\hat {\mathbf{h}}_{i}) 
\end{equation}
 \af{would have the consequence that (under some smoothness assumptions 
 on $f$ and $g$) minimizing the distance between $\mathbf{h}_{i-1}$ and  $\hat{\mathbf{h}}_{i-1}$ should also minimize the loss $L_i$ of the $i$-th layer. 
 This idea is illustrated in the left of Figure \ref{fig:BFAE-diagram}. Indeed,} if the feed-back mappings were the perfect inverses of the feed-forward mappings
($g_i = f_i^{-1}$), one gets
\begin{equation}
 L_i( \hat{\mathbf{h}}_i, f_i(\hat{\mathbf{h}}_{i-1}) ) 
= L_i( \hat{\mathbf{h}}_i, f_i(g_i(\hat{\mathbf{h}}_i)) ) 
= L_i( \hat{\mathbf{h}}_i, \hat{\mathbf{h}}_i )  = 0 \enspace.
\end{equation}
\af{But choosing $g$ to be the perfect inverse of $f$ may need 
heavy computation and
 instability}, since there is no guarantee that $f^{-1}_i$ applied to a target would yield a value that is in the domain of $f_{i-1}$.
\af{An alternative approach is} to learn an approximate inverse
$g_i$, making the $f_i$ / $g_i$ pair look
like an {\em auto-encoder}. This suggests parametrizing
$g_i$ as follows:
\af{
\begin{equation}
 g_i( {\mathbf{h}}_{i}) = \bar s_i( \mathbf{V}_i \mathbf{h}_{i} ), \;\;\; i=1,...,M 
\end{equation}}
where $\bar s_i$ is a non-linearity associated with the decoder and $\mathbf{V}_i$
the matrix of feed-back weights of the $i$-th layer.
With such a parametrization,
it is unlikely that the auto-encoder will achieve zero reconstruction error.
The decoder could be trained via an additional auto-encoder-like loss at each layer
\begin{equation}
  L^{inv}_i 
  = || g_i(f_i({\mathbf{h}}_{i-1}))  - {\mathbf{h}}_{i-1} ||^2_2  \enspace.
\end{equation}
 \af{Changing $\mathbf{V}_i$ based on this loss, makes
 $g$ closer to $f_i^{-1}$. By doing so, it also} makes \af{$ f_i(\hat {\mathbf{h}}_{i-1})=f_i(g_i(\hat{\mathbf{h}}_i))$} closer to
$ \hat{\mathbf{h}}_i $, \af{and is thus also contributing to the decrease of}
$L_i( \hat{\mathbf{h}}_i, f_i(\hat{\mathbf{h}}_{i-1}) ) $.
\af{But we do not want to estimate an inverse mapping only for
the concrete  values we see in training but for a region around the these values to facilitate the computation of $g_i(\hat{\mathbf h_i})$ for $\hat{\mathbf h_i}$ which have never been seen before. For this reason, the loss is modified by noise injection} 
\begin{equation}
\label{eq:target_obj}
L_i^{inv} = || g_i(f_i({\mathbf{h}}_{i-1}+ \epsilon)) 
- ({\mathbf{h}}_{i-1}+ \epsilon) ||^2_2, \;\;\;\; \epsilon \sim N(0,\sigma) \enspace,
\end{equation}
which makes $f_i$ and $g_i$ approximate inverses not just at $\mathbf{h}_{i-1}$
but also in its {\em neighborhood}.

\af{As mentioned above, a required property of target propagation is, that
the layer-wise parameter updates, each improving a layer-wise loss, also lead to an improvement of the global loss. The following theorem shows that, for the case that $g_i$ is a perfect inverse of $f_i$ and  $f_i$ having a certain structure, the update direction of target propagation does not deviate more then 90 degrees from the gradient direction (estimated by back-propagation),  which always leads to a decrease of the global loss.}

\begin{theorem} \footnote{See the proof in the Appendix.}
\textit{Assume that 
 $g_i = f_i^{-1}, i=1,...,M$, and
$f_i$ satisfies
$\mathbf{h}_i=f_i(\mathbf{h}_{i-1}) = W_is_i(\mathbf{h}_{i-1})$ \footnote{This is another way to obtain a non-linear deep network structure.} where $s_i$ can be any differentiable monotonically increasing element-wise function.
Let $\delta W_{i}^{tp}$ and  $\delta W_{i}^{bp}$  be the target propagation update and the back-propagation update in $i$-th layer, respectively. 
If $\hat \eta$  in Equation \eqref{eq:target_toplayer} is sufficiently small, then the angle $\alpha$ between $\delta W_{i}^{tp}$ and $\delta W_{i}^{bp}$ is bounded by
\begin{equation}
0 < \frac{1+\Delta_1(\hat{\eta})}{\frac{\lambda_{max}}{\lambda_{min}}+\Delta_2(\hat\eta)}
\leq cos(\alpha) \leq 1
\end{equation}
Here $\lambda_{max}$ and $\lambda_{min}$ are the largest and smallest singular values
of $(J_{f_{M}}\dots J_{f_{i+1}})^T$, where
$J_{f_k}$ is the Jacobian matrix of $f_k$ and
$\Delta_1(\hat\eta)$ and $\Delta_2(\hat\eta)$ are close to 0 
if $\hat{\eta}$ is sufficiently small.}  
\end{theorem}


\subsection{\af{Difference target propagation}}\label{sec:DiffTargetProp}

\af{
From our experience, the imperfection of the inverse function
leads to severe optimization problems when assigning targets based on equation \eqref{eq:tp}. This  brought us 
to propose the following linearly corrected formula for target
propagation which we refer to as  ``\textit{difference target propagation}'' 
\begin{equation}\label{eq:diff-tp}
\hat {\mathbf{h}}_{i-1}  
=  \mathbf{h}_{i-1}+ g_i(\hat {\mathbf{h}}_{i}) - g_i(\mathbf{h}_{i}) \enspace.
\end{equation}
Note, that if $g_i$ is the inverse of $f_i$,
difference target propagation becomes equivalent to
vanilla target propagation as defined in equation~\eqref{eq:tp}.
The resulting complete training procedure for optimization by difference target propagation is given in Algorithm \ref{alg:experiment-1}}.

\begin{algorithm}[tb] 
\caption{ Training deep neural networks via difference target propagation }
\label{alg:experiment-1}
\begin{algorithmic}


\STATE Compute unit values for all layers:
\FOR {$i=1$ to $M$}
  \STATE $\mathbf{h}_{i} \leftarrow f_i(\mathbf{h}_{i-1})$
\ENDFOR

\STATE Making the first target: $\hat{\mathbf{h}}_{M-1} \leftarrow \mathbf{h}_{M-1} - \hat{\eta} \frac {\partial L} {\partial \mathbf{h}_{M-1}}$, \; ($L$ is the global loss)
\STATE Compute targets for lower layers:
\FOR {$i=M-1$ to $2$}
  \STATE $\hat{\mathbf{h}}_{i-1} \leftarrow \mathbf{h}_{i-1} - g_i(\mathbf{h}_{i}) +  g_i(\hat{\mathbf{h}}_{i})$
\ENDFOR

\STATE Training feedback (inverse) mapping:
\FOR {$i=M-1$ to $2$}
  \STATE Update parameters for $g_i$ using SGD with following a layer-local loss $L_i^{inv}$
  \STATE $L_i^{inv} = ||g_i(f_i(\mathbf{h}_{i-1}+\epsilon)) - (\mathbf{h}_{i-1}+\epsilon) ||_2^2$, \; $\epsilon \sim N(0,\sigma)$
\ENDFOR

\STATE Training feedforward mapping:
\FOR {$i=1$ to $M$}
  \STATE Update parameters for $f_i$ using SGD with following a layer-local loss $L_i$
    \STATE $L_i = ||f_i(\mathbf{h}_{i-1}) - \hat{\mathbf{h}}_{i} ||_2^2$ \; if $i<M$, $L_i = L$ (the global loss) \; if $i=M$.
\ENDFOR

\end{algorithmic}
\end{algorithm}

\af{
In the following, we explain why this linear corrected formula 
stabilizes the optimization process.
In order to achieve stable optimization by target propagation,
$ \mathbf{h}_{i-1} $ should approach  $ \hat{\mathbf{h}}_{i-1} $  
as $ \mathbf{h}_{i} $ approaches  $ \hat{\mathbf{h}}_{i} $. 
Otherwise, the parameters in lower layers continue to be updated
even when an optimum of the global loss is reached already by the upper 
layers, which then could lead the global loss to increase again. 
Thus, the condition 
\begin{equation}
\mathbf{h}_{i} = \hat{\mathbf{h}}_{i} \; \Rightarrow \;
\mathbf{h}_{i-1} = \hat{\mathbf{h}}_{i-1} 
\end{equation}
greatly improves the stability of the optimization.}
This holds for  vanilla  target propagation if  $g_i = f_i^{-1}$, because
\begin{equation}
\mathbf{h}_{i-1} =
f_i^{-1}(\mathbf{h}_{i}) = g_i(\hat{\mathbf{h}}_{i}) =\hat{\mathbf{h}}_{i-1} \enspace.
\end{equation}
\af{Although the condition is not guaranteed to hold for vanilla target propagation if $g_i \neq f_i^{-1}$, for difference target propagation it holds by construction, since}
\begin{equation}
\hat {\mathbf{h}}_{i-1} - \mathbf{h}_{i-1} 
= g_i(\hat {\mathbf{h}}_{i}) - g_i(\mathbf{h}_{i}) \enspace.
\end{equation}

\af{Furthermore, under weak conditions on $f$ and $g$ and if the difference between $\mathbf{h}_{i}$ and $\hat{\mathbf{h}}_{i}$ is small,
we can show for difference target propagation that if the input of the $i$-th layer becomes   $\hat{\mathbf{h}}_{i-1}$ (i.e. the $i-1$-th layer reaches its target)  the output of the $i$-th layer also gets closer to $\hat{\mathbf{h}}_{i}$. This means that the requirement on targets specified by equation \eqref{eq:target_property_local} is met for difference target propagation, as shown in the following theorem}

\begin{theorem}\footnote{See the proof in Appendix.}
Let the target for layer $i-1$  be given by Equation (\ref{eq:diff-tp}), i.e. $\hat {\mathbf{h}}_{i-1}  
=  \mathbf{h}_{i-1}+ g_i(\hat {\mathbf{h}}_{i}) - g_i(\mathbf{h}_{i}) $.
If 
 $ \hat{\mathbf{h}}_i-\mathbf{h}_i$ is sufficiently small, $f_i$  and $g_i$ are differentiable, and
 the corresponding Jacobian matrices $J_{f_i}$ and $J_{g_i}$
satisfy that the largest eigenvalue of 
$ ( I - J_{f_i}J_{g_i})^T ( I - J_{f_i}J_{g_i})$ is less than $1$,
then we have  \begin{equation}
|| \hat{\mathbf{h}}_i - f_i(\hat{\mathbf{h}}_{i-1}) ||^2_2 
< || \hat{\mathbf{h}}_i - \mathbf{h}_i ||^2_2 \enspace.
\end{equation}
\end{theorem}
\zsz{The third condition in the above theorem is easily satisfied in practice, 
because $g_i$ is learned to be the inverse of $f_i$ and makes $g_i\circ f_i$ 
close to the identity mapping, so that $ ( I - J_{f_i}J_{g_i})$ becomes close to the zero matrix which means that the largest
eigenvalue of $ ( I - J_{f_i}J_{g_i})^T ( I - J_{f_i}J_{g_i})$ is also close to $0$.} 
\subsection{\af{Training an auto-encoder with difference target propagation}}\label{sec:diff-tg-ae}

Auto-encoders are interesting \af{ for
learning representations and serve as building blocks for deep neural networks~\citep{Erhan-aistats-2010-small}.}
In addition, as we have seen, training  auto-encoders is part of
\af{the target propagation approach presented here,
where they model the feedback paths used to propagate the targets.}

\af{In the following, we show how a regularized auto-encoder can be trained using difference target propagation instead of back-propagation.}
Like in the work on denoising auto-encoders~\citep{Vincent-JMLR-2010-small} and generative stochastic networks~\citep{Bengio-et-al-ICML2014-small}, we consider the denoising auto-encoder like a stochastic network with noise injected in input
and hidden units, trained to minimize a reconstruction loss. \af{This is, the hidden units are given by the encoder as}
\begin{equation}
\mathbf{h} = f(\mathbf{x}) =
sig(W\mathbf{x}+\mathbf{b})  \enspace,
\end{equation}
\af{where $sig$ is the element-wise sigmoid function, $\mathbf{W}$ the weight matrix and $\mathbf{b}$ the bias vector of the input units. The reconstruction is  given by  the decoder}
\begin{equation}
\mathbf{z} = g(\mathbf{h}) =
sig(W^T(\mathbf{h}+\epsilon)+\mathbf{c}), \;\; 
\epsilon \sim N(0,\sigma)  \enspace ,
\end{equation}
\af{with $\mathbf{c}$ being the bias vector of the hidden units. And the 
reconstruction loss is}
\begin{equation}
L = || \mathbf{z}-\mathbf{x} ||^2_2 + 
|| f(\mathbf{x}+\epsilon)-\mathbf{h} ||^2_2, \;\; 
\epsilon \sim N(0,\sigma)  \enspace ,
\end{equation}
where \af{a regularization term can be added} to obtain a contractive mapping.
In order to train this network without back-propagation (that is, without using the chain rule), we can use difference target propagation as follows
(see  Figure \ref{fig:BFAE-diagram} (right) for an illustration): at first,  the target of $\mathbf{z}$ is just $\mathbf{x}$, so we can train the
reconstruction mapping $g$ based on the loss $ L_g = ||g(\mathbf{h})-\mathbf{x} ||^2_2 $ 
in which $\mathbf{h}$ is considered as a constant. 
Then, we compute the target $\hat{\mathbf{h}}$ of the hidden units following
difference target propagation \af{where we make use of the fact that $f$ is an approximate inverse of $g$.} That is,
\begin{equation}
\hat{\mathbf{h}} = \mathbf{h} + f(\hat{\mathbf{z}}) - f(\mathbf{z}) 
=  2\mathbf{h} - f(\mathbf{z}) \enspace,
\end{equation}
\af{where the last equality follows from $ f(\hat{\mathbf{z}}) = f(\mathbf{x}) = \mathbf{h} $}. As a target loss for 
the hidden layer, we can use $ L_f = || f(\mathbf{x}+\epsilon)-\hat{\mathbf{h}} ||^2_2 $, 
\af{where  $\hat{\mathbf{h}}$ 
is considered as a constant and which can be also augmented by a regularization term to yield a contractive mapping.}

\section{Experiments}\label{sec:experiments}

\af{In a set of experiments we investigated target propagation
for  training deep feedforward deterministic neural networks, networks with discrete transmissions between units, stochastic neural networks, and auto-encoders.}

\af{For training supervised neural networks, we chose the target of the top hidden layer
(number $M-1$) such that it also depends directly on the 
global loss instead of an inverse mapping. That is, we set
$\hat{\mathbf{h}}_{M-1} = \mathbf{h}_{M-1} - \tilde{ \eta} \frac { \partial L (\mathbf{h}_{M}, \mathbf{y} )} { \partial \mathbf{h}_{M-1} } $, where $L$ is the global loss (here the multiclass cross entropy). This may be helpful when the number 
of units in the output layer is much smaller than the number
of units in the top hidden layer, which would make the inverse mapping difficult to learn,
but future work should validate that.}

\dhlee{For discrete stochastic networks in which some form of noise (here Gaussian)
is injected, we used a decaying noise level  for learning the inverse mapping,
in order to stabilize learning, i.e. the
standard deviation of the Gaussian is set to 
$\sigma(e) = \sigma_0 / ( 1 + e/e_0 ) $ where $\sigma_0$ is the initial value, $e$ is the 
epoch number and $e_0$ is the half-life of this decay. 
This seems to help to  fine-tune the feedback weights at the end of training.}

\af{In all experiments, the weights were initialized with orthogonal random matrices and the bias parameters were initially set to zero.  All experiments were repeated 10 times with different random initializations.} We put the code of these experiments online (\url{https://github.com/donghyunlee/dtp}).

\subsection{Deterministic feedforward deep networks}

As a primary objective, we investigated training of ordinary deep supervised
networks with continuous and deterministic units on the MNIST dataset. We used a held-out validation set of 10000 samples for choosing hyper-parameters.  
We trained networks with 7 hidden layers each consisting of 240 units (using the hyperbolic
tangent as activation function) with difference target propagation and 
back-propagation.

Training was based on RMSprop~\citep{tieleman2012lecture} 
where hyper-parameters for the best validation error were found using random search~\citep{Bergstra+Bengio-2012-small}. RMSprop is an adaptive learning rate algorithm known to lead to good results for back-propagation. Furthermore, it is suitable for updating the parameters of each layer based on the layer-wise
targets obtained by target propagation. 
Our experiments suggested that when using a hand-selected learning
rate per layer rather than the automatically set one (by RMSprop), the selected learning rates 
were different for each layer, which is why we decided to use an adaptive
method like RMSprop.


\begin{figure}[htp]
\vspace{-10pt}
\begin{center}
\centerline{\includegraphics[width=0.5\columnwidth]{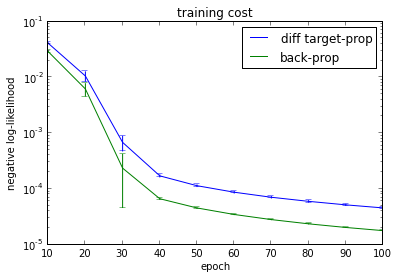}\includegraphics[width=0.5\columnwidth]{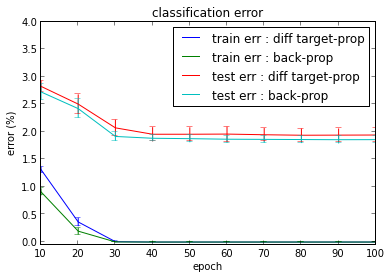}}
\caption{Mean training cost (left) and train/test classification error (right) with target propagation
and back-propagation using continuous deep networks (tanh) on MNIST. 
Error bars indicate the standard deviation. 
}
\label{fig:mnist_7h_tanh}
\end{center}
\vspace{-30pt}
\end{figure}

\af{The results are shown in Figure \ref{fig:mnist_7h_tanh}. We  
obtained a test error of 1.94\%  with target propagation and 1.86\% with back propagation. 
The final negative log-likelihood on the training set was $4.584 \times 10^{-5}$
with target propagation  and $1.797 \times 10 ^ {-5}$ with back propagation. 
We also trained the same network with
rectifier linear units and got a test error of 3.15\%  whereas
 1.62\%  was obtained with back-propagation. It is 
 well known that this nonlinearity is advantageous for back-propagation,
while it  seemed to be less appropriate for this implementation
of target propagation.}

\af{In a second experiment we investigated training on CIFAR-10.
The experimental setting was the same as for MNIST (using the hyperbolic tangent as activation function) except that the network architecture
was 3072-1000-1000-1000-10. We did not use any preprocessing, except
for scaling the input values to lay in [0,1], 
and we tuned the hyper-parameters of RMSprop using a held-out validation set of 1000  samples.
We obtained mean test accuracies of 50.71\%  and 53.72\% 
for target propagation and  back-propagation, respectively. 
It was reported in \cite{Krizhevsky-2009}, that a network with 1 hidden layer of 1000 units 
achieved 49.78\% accuracy with back-propagation, and increasing the number of units to 10000  led to 51.53\% accuracy. 
As the current state-of-the-art performance on the permutation
invariant CIFAR-10 recognition task, ~\cite{kishore-2015} reported 64.1\% but when using PCA without whitening as preprocessing and 
zero-biased auto-encoders for unsupervised pre-training.}

\subsection{Networks with discretized transmission between units}

\begin{figure}[ht]
\vspace{-10pt}
\begin{center}
\includegraphics[width=0.5\columnwidth]{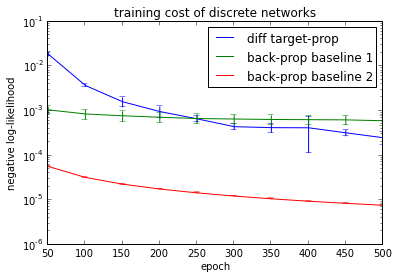}\includegraphics[width=0.5\columnwidth]{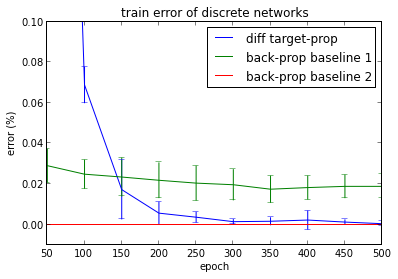}
\includegraphics[width=0.5\columnwidth]{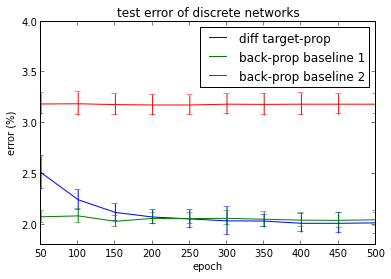}\hfill \includegraphics[width=0.4\columnwidth]{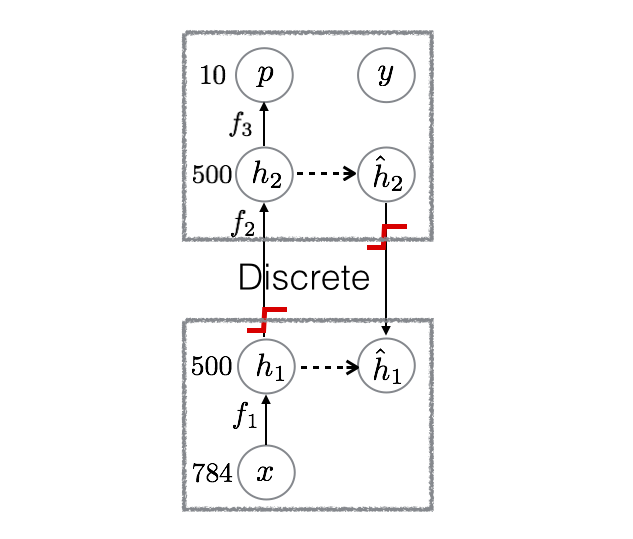}
\caption{
Mean training cost (top left), mean training error (top right)
and mean test error (bottom left) while training discrete networks with difference target propagation and the two baseline versions of back-propagation. Error bars indicate standard deviations over the 10 runs.\,\,
Diagram of the discrete network (bottom right). 
The output of $\textbf{h}_1$ is discretized 
because signals must be communicated from $\textbf{h}_1$ to $\textbf{h}_2$ through a long cable, so binary representations are preferred in order to conserve energy.
With target propagation, training signals are also discretized through this cable 
(since feedback paths are computed by bona-fide neurons).
}
\label{fig:mnist_disc_test}
\end{center}
\vspace{-20pt}
\end{figure}

\af{To explore target propagation for an extremely non-linear neural network,
we investigated training of discrete networks on the MNIST dataset.
The network architecture was 784-500-500-10,
where only the 1st hidden layer was discretized.
Inspired by biological considerations and the objective of reducing
the communication cost between neurons, 
instead of just using the step activation function, we used 
ordinary neural net layers but 
with signals being discretized when transported between the first and second layer. 
The network structure is depicted in 
the right plot of Figure \ref{fig:mnist_disc_test} and the activations of the hidden layers are given by}
\begin{equation}
\mathbf{h}_1 = f_1(\mathbf{x}) 
= \tanh( W_1 \mathbf{x} )  \enspace \,\,\, \text{and} \enspace \,\,\, \mathbf{h}_2 
= f_2(\mathbf{h}_{1}) = \tanh( W_2 sign(\mathbf{h}_{1}) )
\end{equation}
where $ sign(x) = 1 $ if $ x > 0 $, and $sign(x)=0$ if $ x \leq 0 $. 
The network output is given by
\begin{equation}
p(\mathbf{y}|\mathbf{x}) = f_3(\mathbf{h}_{2}) = 
softmax( W_3 \mathbf{h}_{2} )  \enspace.
\end{equation}
\af{
The inverse mapping of the second layer and the associated 
loss are given by 
}
\begin{equation}
g_2(\mathbf{h}_{2}) = \tanh( V_2 sign(\mathbf{h}_{2}) )  \enspace,
\end{equation}
\begin{equation}
L_2^{inv} = || g_2(f_2(\mathbf{h}_{1}+ \epsilon)) 
- (\mathbf{h}_{1}+ \epsilon) ||^2_2, \;\;\;\; \epsilon \sim N(0,\sigma) 
\enspace.
\end{equation}
\af{If feed-forward mapping is discrete, 
back-propagated gradients become 0 and useless when they
cross the discretization step.
So we compare target propagation to two baselines.
As a first baseline, we train the network with back-propagation 
and the {\em straight-through estimator}~\citep{bengio2013estimating},
which is biased but was found to work well, and
simply ignores the derivative of the step function (which is 0 or
infinite) in the back-propagation phase.
As a second baseline, we train only the upper layers by back-propagation,
while not changing the weight $W_1$ which are affected by
the discretization, i.e., the lower layers do not learn.}

\af{The results  on the training and test sets are shown in Figure 
\ref{fig:mnist_disc_test}. 
The training error for the first baseline (straight-through estimator)
does not converge to zero
(which can be explained by the biased gradient) but generalization performance is fairly good.
The second baseline (fixed lower layer) surprisingly
reached zero training error, but did not perform well 
on the test set. This can be explained by the fact that
it cannot learn any meaningful representation at the first layer.
Target propagation however did not suffer from this drawback and 
can be used to train discrete networks directly 
(training signals can pass the discrete region successfully). 
Though the training convergence was slower,  the training error 
did approach zero. In addition, difference target propagation also 
achieved good results on the test set.
}


\subsection{Stochastic networks}

\af{Another interesting model class which vanilla back-propagation cannot  deal with are stochastic networks with discrete units. Recently, stochastic networks have attracted
attention~\citep{Bengio-arxiv2013,Tang+Salakhutdinov-2013,bengio2013estimating} 
because they are able to learn  a multi-modal conditional
distribution $P(Y|X)$, which
is important for structured output predictions.
Training networks of stochastic binary units is
also biologically motivated, since they resemble networks of spiking neurons.
Here, we investigate whether one can train networks of
stochastic binary units on MNIST for classification
using target propagation.}
Following~\cite{Raiko-et-al-NIPS-dlworkshop2014}, the network architecture was 784-200-200-10
and the hidden units were stochastic binary units with the 
probability of turning on given by a sigmoid activation:
\af{\begin{equation}
\mathbf{h}_{i}^p 
= P(\mathbf{H}_i=1| \mathbf{h}_{i-1})
= \sigma(W_i\mathbf{h}_{i-1}), \;\;\; 
\mathbf{h}_{i} \sim P(\mathbf{H}_i| \mathbf{h}_{i-1}) \enspace,
\end{equation}}
\af{that is, $\mathbf h_i$  is one with probability $\mathbf{h}_{i}^p $.}

As a baseline, we considered training based on the {\em straight-through} 
biased gradient estimator~\citep{bengio2013estimating} in which the
derivative through the discrete sampling step is ignored
(this method showed the best performance in~\cite{Raiko-et-al-NIPS-dlworkshop2014}.)
That is
\begin{equation}
\delta \mathbf{h}_{i-1}^p = \delta \mathbf{h}_{i}^p \frac{\partial
\mathbf{h}_{i}^p} {\partial \mathbf{h}_{i-1}^p} 
\approx \sigma'(W_i\mathbf{h}_{i-1}) W_i^T 
\delta \mathbf{h}_{i}^p  \enspace.
\end{equation}
\af{
With difference  target propagation the stochastic network can be trained directly, setting the targets to}
\begin{equation}
\hat{\mathbf{h}}_{2}^p = \mathbf{h}_{2}^p - 
\eta \frac { \partial L  } { \partial \mathbf{h}_{2}  } \;\;\;\;\;\; \text{and} \;\;\;\;\;\;
\hat{\mathbf{h}}_{1}^p = \mathbf{h}_{1}^p + g_2(\hat{\mathbf{h}}_{2}^p)
- g_2(\mathbf{h}_{2}^p) 
\end{equation}
\af{where $g_i(\mathbf{h}_i^p) = \tanh(V_i \mathbf{h}_i^p)$ is trained by the loss} 
\begin{equation}
L_i^{inv} = || g_i(f_i(\mathbf{h}_{i-1}+ \epsilon)) 
- (\mathbf{h}_{i-1}+ \epsilon) ||^2_2, \;\;\;\; \epsilon \sim N(0,\sigma)  \enspace,
\end{equation}
and  layer-local target losses are defined as $ L_i = || \hat{\mathbf{h}}_i^p -
\mathbf{h}_i^p ||^2_2 $. 

\begin{table}[ht]
\centering 
\begin{tabular}{c c c} 
\hline\hline 
Method & Test Error(\%) \\ [0.5ex] 
\hline 
Difference Target-Propagation, M=1 & 1.54\%  \\ [0.5ex]
Straight-through gradient estimator \citep{bengio2013estimating} + backprop, M=1\\ 
as reported in \citet{Raiko-et-al-NIPS-dlworkshop2014} & 1.71\% \\ [0.5ex]
\hline 
as reported in \citet{Tang+Salakhutdinov-2013}, M=20 & 3.99\% \\ [0.5ex] 
as reported in \citet{Raiko-et-al-NIPS-dlworkshop2014}, M=20 & 1.63\% \\ [0.5ex] 
\hline 
\end{tabular}
\vspace{4mm}
\caption{Mean test Error on MNIST for stochastoc networks. The first row shows
the results of our experiments averaged  over 10 trials.
The second row shows the results reported in~\citep{Raiko-et-al-NIPS-dlworkshop2014}.
M corresponds to the number of samples used  for computing output probabilities. We used  M=1 during training and M=100 for the test set.} 
\label{table:nonlin} 
\end{table}

\af{For evaluation, we averaged the output probabilities for a given input over 100  samples, and  classified the example accordingly, 
following~\cite{Raiko-et-al-NIPS-dlworkshop2014}.
Results are given in Table \ref{table:nonlin}.
We obtained a test error of 1.71\% using the baseline method
and 1.54\% using target propagation, which is -- to our knowledge --  the best result for stochastic nets on MNIST reported so far. This suggests that target propagation is highly promising for training networks of binary stochastic units.}

\subsection{\af{Auto-encoder}}

\begin{figure}
\vspace{-10pt}
\begin{center}
\centerline{
\includegraphics[width=0.5\columnwidth]{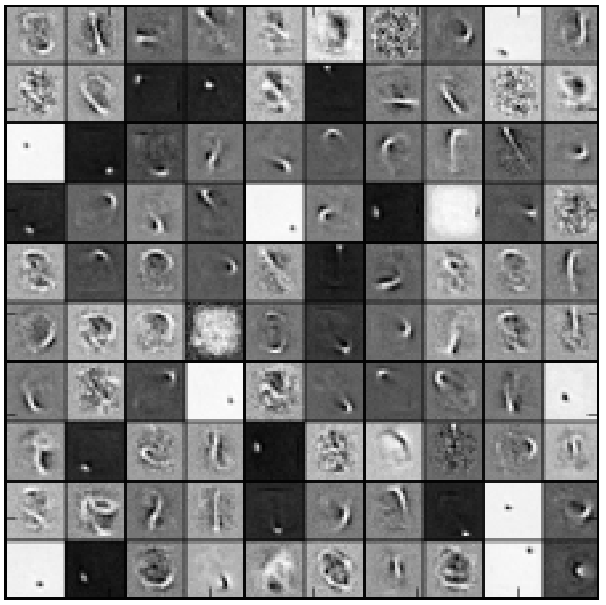}}
\caption{Filters learned by the back-propagation-free auto-encoder. Each filter corresponds to the hidden weights of 
one of 100 randomly chosen hidden units. We obtain stroke filters, similar to those usually obtained by regularized auto-encoders.}
\label{fig:BFAE}
\end{center}
\vspace{-30pt}
\end{figure}

\af{We trained a denoising auto-encoder with 1000 hidden units with 
difference target propagation as described in Section \ref{sec:diff-tg-ae} on MNIST.
As shown in Figure \ref{fig:BFAE} stroke-like filters can be obtained by target propagation. After supervised fine-tuning (using back-propagation), we got a test error of 1.35$\%$.
Thus, by training an auto-encoder with target propagation one can learn a good initial representation, which is as good as the one obtained by  regularized auto-encoders trained by back-propagation on the reconstruction error.}

\section{Conclusion}
\af{We introduced  a novel optimization method for neural networks, called 
target propagation, which was designed to overcome drawbacks
of  back-propagation and is biologically more plausible.  Target propagation replaces training signals based on partial derivatives by targets
which are propagated based on an auto-encoding feedback loop. Difference target propagation is a linear correction for this imperfect inverse mapping which is  effective to make target propagation actually work.
 Our experiments show that target propagation performs  comparable to back-propagation on ordinary deep networks and denoising auto-encoders. Moreover, target propagation can be directly used on networks with discretized transmission between units and reaches state of the art performance for stochastic neural networks on MNIST. 
}

\medskip

\begin{small}
\paragraph{\textbf{Acknowledgments}}
We would like to thank Junyoung Chung for providing RMSprop code, 
Caglar Gulcehre and Antoine Biard for general discussion and feedback, 
Jyri Kivinen for discussion of backprop-free auto-encoder,
Mathias Berglund for explanation of his stochastic networks. 
We thank the developers of Theano ~\citep{bergstra+al:2010-scipy,Bastien-Theano-2012}, 
a Python library which allowed us to easily develop a fast and optimized code for GPU.
We are also grateful for funding from NSERC, 
the Canada Research Chairs, Compute Canada, and CIFAR.
\end{small}

\bibliographystyle{splncsnat}

\bibliography{strings,ml,aigaion,dtp}

\begin{thebibliography}{22}
\providecommand{\natexlab}[1]{#1}
\providecommand{\url}[1]{\texttt{#1}}
\providecommand{\urlprefix}{}

\bibitem[{Bastien et~al.(2012)Bastien, Lamblin, Pascanu, Bergstra, Goodfellow,
  Bergeron, Bouchard, and Bengio}]{Bastien-Theano-2012}
Bastien, F., Lamblin, P., Pascanu, R., Bergstra, J., Goodfellow, I.J.,
  Bergeron, A., Bouchard, N., Bengio, Y.: Theano: new features and speed
  improvements.
\newblock Deep Learning and Unsupervised Feature Learning NIPS 2012 Workshop
  (2012)

\bibitem[{Bengio(2009)}]{Bengio-2009-book}
Bengio, Y.: Learning deep architectures for {AI}.
\newblock Now Publishers (2009)

\bibitem[{Bengio(2013)}]{Bengio-arxiv2013}
Bengio, Y.: Estimating or propagating gradients through stochastic neurons.
\newblock Tech. Rep. arXiv:1305.2982, Universite de Montreal (2013)

\bibitem[{Bengio(2014)}]{Bengio-arxiv2014}
Bengio, Y.: How auto-encoders could provide credit assignment in deep networks
  via target propagation.
\newblock Tech. rep., arXiv:1407.7906 (2014)

\bibitem[{Bengio et~al.(2013)Bengio, L{\'e}onard, and
  Courville}]{bengio2013estimating}
Bengio, Y., L{\'e}onard, N., Courville, A.: Estimating or propagating gradients
  through stochastic neurons for conditional computation.
\newblock arXiv:1308.3432 (2013)

\bibitem[{Bengio et~al.(2014)Bengio, Thibodeau-Laufer, and
  Yosinski}]{Bengio-et-al-ICML2014-small}
Bengio, Y., Thibodeau-Laufer, E., Yosinski, J.: Deep generative stochastic
  networks trainable by backprop.
\newblock In: ICML'2014 (2014)

\bibitem[{Bergstra and Bengio(2012)}]{Bergstra+Bengio-2012-small}
Bergstra, J., Bengio, Y.: Random search for hyper-parameter optimization.
\newblock J. Machine Learning Res. 13, 281--305 (2012)

\bibitem[{Bergstra et~al.(2010)Bergstra, Breuleux, Bastien, Lamblin, Pascanu,
  Desjardins, Turian, Warde-Farley, and Bengio}]{bergstra+al:2010-scipy}
Bergstra, J., Breuleux, O., Bastien, F., Lamblin, P., Pascanu, R., Desjardins,
  G., Turian, J., Warde-Farley, D., Bengio, Y.: Theano: a {CPU} and {GPU} math
  expression compiler.
\newblock In: Proceedings of the Python for Scientific Computing Conference
  ({SciPy}) (Jun 2010), oral Presentation

\bibitem[{Carreira-Perpinan and
  Wang(2014)}]{Carreira-Perpinan-and-Wang-AISTATS2014}
Carreira-Perpinan, M., Wang, W.: Distributed optimization of deeply nested
  systems.
\newblock In: AISTATS'2014, JMLR W\&CP. vol.~33, pp. 10--19 (2014)

\bibitem[{Erhan et~al.(2010)Erhan, Courville, Bengio, and
  Vincent}]{Erhan-aistats-2010-small}
Erhan, D., Courville, A., Bengio, Y., Vincent, P.: Why does unsupervised
  pre-training help deep learning?
\newblock In: JMLR W\&CP: Proc. AISTATS'2010. vol.~9, pp. 201--208 (2010)

\bibitem[{Hinton et~al.(2012)Hinton, Deng, Dahl, Mohamed, Jaitly, Senior,
  Vanhoucke, Nguyen, Sainath, and Kingsbury}]{Hinton-et-al-2012}
Hinton, G., Deng, L., Dahl, G.E., Mohamed, A., Jaitly, N., Senior, A.,
  Vanhoucke, V., Nguyen, P., Sainath, T., Kingsbury, B.: Deep neural networks
  for acoustic modeling in speech recognition.
\newblock {IEEE} Signal Processing Magazine 29(6), 82--97 (Nov 2012)

\bibitem[{Konda et~al.(2015)Konda, Memisevic, and Krueger}]{kishore-2015}
Konda, K., Memisevic, R., Krueger, D.: Zero-bias autoencoders and the benefits
  of co-adapting features.
\newblock Under review on International Conference on Learning Representations
  (2015)

\bibitem[{Krizhevsky et~al.(2012)Krizhevsky, Sutskever, and
  Hinton}]{Krizhevsky-2012-small}
Krizhevsky, A., Sutskever, I., Hinton, G.: {ImageNet} classification with deep
  convolutional neural networks.
\newblock In: NIPS'2012 (2012)

\bibitem[{Krizhevsky and Hinton(2009)}]{Krizhevsky-2009}
Krizhevsky, A., Hinton, G.: Learning multiple layers of features from tiny
  images.
\newblock Master's thesis, University of Toronto (2009)

\bibitem[{{LeCun}(1986)}]{LeCun-dsbo86}
{LeCun}, Y.: Learning processes in an asymmetric threshold network.
\newblock In: Fogelman-Souli\'e, F., Bienenstock, E., Weisbuch, G. (eds.)
  Disordered Systems and Biological Organization, pp. 233--240.
  Springer-Verlag, Les Houches, France (1986)

\bibitem[{{LeCun}(1987)}]{Lecun-these87}
{LeCun}, Y.: Mod\`eles connexionistes de l'apprentissage.
\newblock Ph.D. thesis, Universit\'e de Paris VI (1987)

\bibitem[{Raiko et~al.(2014)Raiko, Berglund, Alain, and
  Dinh}]{Raiko-et-al-NIPS-dlworkshop2014}
Raiko, T., Berglund, M., Alain, G., Dinh, L.: Techniques for learning binary
  stochastic feedforward neural networks.
\newblock NIPS Deep Learning Workshop 2014  (2014)

\bibitem[{Sutskever et~al.(2014)Sutskever, Vinyals, and
  Le}]{Sutskever-et-al-arxiv2014}
Sutskever, I., Vinyals, O., Le, Q.V.: Sequence to sequence learning with neural
  networks.
\newblock Tech. rep., arXiv:1409.3215 (2014)

\bibitem[{Szegedy et~al.(2014)Szegedy, Liu, Jia, Sermanet, Reed, Anguelov,
  Erhan, Vanhoucke, and Rabinovich}]{Szegedy-et-al-arxiv2014}
Szegedy, C., Liu, W., Jia, Y., Sermanet, P., Reed, S., Anguelov, D., Erhan, D.,
  Vanhoucke, V., Rabinovich, A.: Going deeper with convolutions.
\newblock Tech. rep., arXiv:1409.4842 (2014)

\bibitem[{Tang and Salakhutdinov(2013)}]{Tang+Salakhutdinov-2013}
Tang, Y., Salakhutdinov, R.: A new learning algorithm for stochastic
  feedforward neural nets.
\newblock ICML'2013 Workshop on Challenges in Representation Learning (2013)

\bibitem[{Tieleman and Hinton(2012)}]{tieleman2012lecture}
Tieleman, T., Hinton, G.: Lecture 6.5-rmsprop: Divide the gradient by a running
  average of its recent magnitude.
\newblock COURSERA: Neural Networks for Machine Learning 4 (2012)

\bibitem[{Vincent et~al.(2010)Vincent, Larochelle, Lajoie, Bengio, and
  Manzagol}]{Vincent-JMLR-2010-small}
Vincent, P., Larochelle, H., Lajoie, I., Bengio, Y., Manzagol, P.A.: Stacked
  denoising autoencoders: Learning useful representations in a deep network
  with a local denoising criterion.
\newblock J. Machine Learning Res. 11 (2010)

\end{thebibliography}
\appendix

\renewcommand{\theequation}{A-\arabic{equation}}
\setcounter{equation}{0}  
\section*{\huge Appendix}  
\section{Proof of Theorem 1} \label{app:app1}  
\begin{proof}
\af{Given a training example $(\mathbf{x}, \mathbf{y})$ the back-propagation update is given by}
\begin{equation}
\nonumber \delta W_{i}^{bp} = -\frac{\partial L(\mathbf{x}, \mathbf{y};\mathbf{\theta}_W^{0,M})}{\partial W_i} = -J_{f_{i+1}}^T \dots J_{f_{M}}^T \frac{\partial L}{\partial \mathbf{h}_{M}}(s_i(\mathbf{h}_{i-1}))^T \enspace,
\end{equation}
where
$J_{f_k}=\frac{\partial \mathbf{h}_{k}}{\partial \mathbf{h}_{k-1}} = W_i \cdot S_i'(\mathbf{h}_{k-1}), k=i+1,\dots,M$. \af{Here}
$S_i'(\mathbf{h}_{k-1})$ is  a diagonal \af{matrix}
with each diagonal element being element-wise derivatives
and $J_{f_k}$ is the Jacobian of $f_k(\mathbf{h}_{k-1})$.
In target propagation \af{the target for 
 $\mathbf{h}_{M}$ is given by}
$\hat{\mathbf{h}}_{M} = \mathbf{h}_{M} - \hat{\eta}\frac{\partial L}{\partial \mathbf{h}_{M}}$.
If all $\mathbf{h}_k$'s are allocated in smooth areas and $\hat{\eta}$ is sufficiently small,
\af{ we can apply a Taylor expansion to get}
\begin{equation}
\nonumber \hat{\mathbf{h}}_i = g_{i+1}(\dots g_{M}(\hat{\mathbf{h}}_{M})\dots) =  g_{i+1}(\dots g_{M}(\mathbf{h}_{M})\dots) - \hat{\eta} J_{g_{i+1}}\dots J_{g_{M}}\frac{\partial L}{\partial \mathbf{h}_{M}}+\mathbf{o}(\hat{\eta}) \enspace,
\end{equation}
\af{where} $\mathbf{o}(\hat{\eta})$ is the remainder satisfying $\lim_{\hat{\eta} \rightarrow 0} \mathbf{o}(\hat{\eta})/\hat{\eta} = \mathbf{0}$.
 Now, for $\delta W_{i}^{tp}$ we have
\begin{eqnarray}
\nonumber \delta W_{i}^{tp} &=& -\frac{\partial||\mathbf{h}_i(\mathbf{h}_{i-1};W_i)-\hat{\mathbf{h}}_i||^2_2}{\partial W_i} \\
\nonumber &=& - (\mathbf{h}_i-(\mathbf{h}_i-\hat{\eta} J_{f_{i+1}}^{-1}\dots J_{f_{M}}^{-1}\frac{\partial L}{\partial \mathbf{h}_{M}}+\mathbf{o}(\hat{\eta})))(s_i(\mathbf{h}_{i-1}))^T\\ 
\nonumber &=& - \hat{\eta}J_{f_{i+1}}^{-1}\dots J_{f_{M}}^{-1}\frac{\partial L}{\partial \mathbf{h}_{M}}(s_i(\mathbf{h}_{i-1}))^T
+\mathbf{o}(\hat{\eta}) (s_i(\mathbf{h}_{i-1}))^T \enspace.
\end{eqnarray}
We write $\frac{\partial L}{\partial \mathbf{h}_{M}}$ as $\textit{\textbf{l}}$
, $s_i(\mathbf{h}_{i-1})$ as \textit{\textbf{v}} and $J_{f_{M}}\dots J_{f_{i+1}}$ as $J$ for short.
Then the inner production of vector forms of $\delta W_{i}^{bp}$ and $\delta W_{i}^{tp}$ is
\begin{eqnarray}
\nonumber &&\langle vec(\delta W_{i}^{bp}), vec(\delta W_{i}^{tp}) \rangle
= tr((J^T \textit{\textbf{l}}\textit{\textbf{v}}^T)^T(\hat{\eta}J^{-1} \textit{\textbf{l}}\textit{\textbf{v}}^T + \mathbf{o}(\hat{\eta}) \textit{\textbf{v}}^T)) \\ 
\nonumber &=& \hat{\eta}tr(\textit{\textbf{v}}\textit{\textbf{l}}^TJJ^{-1} \textit{\textbf{l}}\textit{\textbf{v}}^T)
- tr(\textit{\textbf{v}}\textit{\textbf{l}}^TJ\mathbf{o}(\hat{\eta})\textit{\textbf{v}}^T)
=\hat{\eta}||\textit{\textbf{v}}||^2_2||\textit{\textbf{l}}||^2_2
- \langle J^T\textit{\textbf{l}}, \mathbf{o}(\hat{\eta})\rangle||\textit{\textbf{v}}||^2_2 \enspace.
\end{eqnarray}
For $||vec(\delta W_{i}^{bp})||_2$ and $||vec(\delta W_{i}^{tp})||_2$ we have
\begin{eqnarray}
\nonumber && ||vec(\delta W_{i}^{bp})||_2 
 = \sqrt{tr((-J^T \textit{\textbf{l}}\textit{\textbf{v}}^T)^T(-J^T \textit{\textbf{l}}\textit{\textbf{v}}^T))}
=||\textit{\textbf{v}}||_2||J^T \textit{\textbf{l}}||_2
\leq ||\textit{\textbf{v}}||_2||J^T||_2||\textit{\textbf{l}}||_2
\end{eqnarray}
and similarly
\begin{equation}
\nonumber ||vec(\delta W_{i}^{tp})||_2 
\leq \hat{\eta}||\textit{\textbf{v}}||_2||J^{-1}||_2||\textit{\textbf{l}}||_2
+||\mathbf{o}(\hat{\eta})||_2||\textit{\textbf{v}}||_2 \enspace,
\end{equation}
\af{where} $||J^T||_2$ and $||J^{-1}||_2$ are matrix Euclidean norms,
i.e. the largest singular value of $(J_{f_{M}}\dots J_{f_{i+1}})^T$, $\lambda_{max}$,
				and the largest singular value of $(J_{f_{M}}\dots J_{f_{i+1}})^{-1}$, $\frac{1}{\lambda_{min}}$
				($\lambda_{min}$ is the smallest singular value of $(J_{f_{M}}\dots J_{f_{i+1}})^T$,
				 because $f_k$ is invertable, so all the smallest singular values of Jacobians are larger than $0$).
Finally, if $\hat{\eta}$ is sufficiently small, the angle $\alpha$ between $vec(\delta W_{i}^{bp})$ and $vec(\delta W_{i}^{tp})$ satisfies:
\begin{eqnarray}
\nonumber cos(\alpha) &=& \frac{\langle vec(\delta W_{i}^{bp}), vec(\delta W_{i}^{tp}) \rangle}{||vec(\delta W_{i}^{bp})||_2\cdot||vec(\delta W_{i}^{tp})||_2}\\
\nonumber &\geq& \frac{\hat{\eta}||\textit{\textbf{v}}||^2_2||\textit{\textbf{l}}||^2_2
- \langle J^T\textit{\textbf{l}}, \mathbf{o}(\hat{\eta})\rangle||\textit{\textbf{v}}||^2_2}
{(||\textit{\textbf{v}}||_2\lambda_{max}||\textit{\textbf{l}}||_2)
(\hat{\eta}||\textit{\textbf{v}}||_2(\frac{1}{\lambda_{min}})||\textit{\textbf{l}}||_2
+||\mathbf{o}(\hat{\eta})||_2||\textit{\textbf{v}}||_2)}\\
\nonumber &=& \frac{1+\frac{- \langle J^T\textit{\textbf{l}}, \mathbf{o}(\hat{\eta})\rangle}{\hat{\eta}||\textit{\textbf{l}}||^2_2}}
{\frac{\lambda_{max}}{\lambda_{min}}+\frac{\lambda_{max}||\mathbf{o}(\hat{\eta})||_2}{\hat{\eta}||\textit{\textbf{l}}||_2}}
= \frac{1+\Delta_1(\hat{\eta})}
{\frac{\lambda_{max}}{\lambda_{min}}+\Delta_2(\hat{\eta})}
\end{eqnarray}
\af{where the last expression is positive if $\hat\eta$ is sufficiently small} and $cos(\alpha)\leq 1$ is trivial.
\end{proof}
\section{Proof of Theorem 2} \label{app:app2}  
\begin{proof}
Let $\mathbf{e} =\hat{\mathbf{h}}_i -\mathbf{h}_i $.
\af{Applying Taylor's theorem twice, we get}
\begin{eqnarray}
\nonumber \hat{\mathbf{h}}_i - f_i(\hat{\mathbf{h}}_{i-1}) &=&
 \hat{\mathbf{h}}_i - f_i(\mathbf{h}_{i-1} + g_i(\hat {\mathbf{h}}_{i}) - g_i(\mathbf{h}_{i}) )
=  \hat{\mathbf{h}}_i - f_i(\mathbf{h}_{i-1} + J_{g_i}\mathbf{e} + \mathbf{o}(||\mathbf{e}||_2)) \\ 
\nonumber &=&  \hat{\mathbf{h}}_i - f_i(\mathbf{h}_{i-1}) - J_{f_i}(J_{g_i}\mathbf{e} + \mathbf{o}(||\mathbf{e}||_2)) 
- \mathbf{o}(|| J_{g_i}\mathbf{e} + \mathbf{o}(||\mathbf{e}||_2)||_2)\\
\nonumber&=&  \hat{\mathbf{h}}_i - \mathbf{h}_{i} - J_{f_i}J_{g_i}\mathbf{e} - \mathbf{o}(||\mathbf{e}||_2)
= (I - J_{f_i}J_{g_i})\mathbf{e} - \mathbf{o}(||\mathbf{e}||_2) 
\end{eqnarray}
\af{where  the vector} $\mathbf{o}(||\mathbf{e}||_2)$ represents the remainder satisfying $\lim_{\mathbf{e}\rightarrow \mathbf{0}} \mathbf{o}(||\mathbf{e}||_2)/||\mathbf{e}||_2 = \mathbf{0}$.
Then for $|| \hat{\mathbf{h}}_i - f_i(\hat{\mathbf{h}}_{i-1}) ||^2_2 $
we have  
\begin{eqnarray}
\nonumber || \hat{\mathbf{h}}_i - f_i(\hat{\mathbf{h}}_{i-1}) ||^2_2
\nonumber&=& ((I - J_{f_i}J_{g_i})\mathbf{e} - \mathbf{o}(||\mathbf{e}||_2))^T
((I - J_{f_i}J_{g_i})\mathbf{e} - \mathbf{o}(||\mathbf{e}||_2)) \\
\nonumber&=&  \mathbf{e}^T(I - J_{f_i}J_{g_i})^T(I - J_{f_i}J_{g_i})\mathbf{e} - \mathbf{o}(||\mathbf{e}||_2)^T
(I - J_{f_i}J_{g_i})\mathbf{e}\\
\nonumber &&- \mathbf{e}^T(I - J_{f_i}J_{g_i})^T\mathbf{o}(||\mathbf{e}||_2) +
\mathbf{o}(||\mathbf{e}||_2)^T\mathbf{o}(||\mathbf{e}||_2)) \\
\nonumber&=& \mathbf{e}^T(I - J_{f_i}J_{g_i})^T(I - J_{f_i}J_{g_i})\mathbf{e} + o(||\mathbf{e}||^2_2)\\
\label{eq:th2_eq2}
&\leq& \lambda||\mathbf{e}||^2_2 + |o(||\mathbf{e}||^2_2)|
\end{eqnarray}
where  $o(||\mathbf{e}||^2_2)$ is the scalar value resulting from all terms depending on $\mathbf{o}(||\mathbf{e}||_2)$ and $\lambda$ is the 
largest eigenvalue of 
$ ( I - J_{f_i}J_{g_i})^T ( I - J_{f_i}J_{g_i})$.
If $\mathbf{e}$ is sufficiently small to guarantee 
$|o(||\mathbf{e}||^2_2)|< (1-\lambda)||\mathbf{e}||^2_2$,
then the left of Equation (\ref{eq:th2_eq2}) is less than $||\mathbf{e}||^2_2$ which is just
$|| \hat{\mathbf{h}}_i - \mathbf{h}_i ||^2_2$.\\

\end{proof}


\end{document}